\documentclass[runningheads]{llncs}
\usepackage{graphicx}
\usepackage{cite} 
\usepackage{times}
\usepackage{epsfig}
\usepackage{amsmath}
\usepackage{amssymb}
\usepackage{mwe}
\usepackage{acro}
\usepackage{amssymb}
\usepackage{xcolor,colortbl}
\usepackage{tabularx}
\usepackage{relsize}
\usepackage{pifont}
\usepackage{booktabs} 
\usepackage{multirow}
\usepackage{multicol}
\usepackage{adjustbox}
\usepackage{float}
\usepackage[colorlinks,
            linkcolor=red,  
            anchorcolor=blue, 
            citecolor=green,   
            ]{hyperref}
\usepackage{amsmath}

\begin{document}

\title{Survival Prediction of Brain Cancer with Incomplete Radiology, Pathology, Genomic, and Demographic Data}
%index{Last Name, First Name}
\author{Can Cui\inst{1} \and
Han Liu\inst{1} \and
Quan Liu\inst{1} \and
Ruining Deng\inst{1} \and
Zuhayr Asad\inst{1} \and
Yaohong Wang\inst{2} \and
Shilin Zhao\inst{2} \and
Haichun Yang\inst{2} \and
Bennett A. Landman\inst{1} \and
Yuankai Huo\inst{1}}
%\author{Cui, Can; Liu, Han; Liu, Quan; Deng, Ruining; Asad, Zuhayr; Wang, Yaohong;  Zhao, Shilin; Yang, Haichun; Landman, Bennett A.; Huo, Yuankai} 

\institute{
Vanderbilt University, Nashville TN 37235, USA \and
Vanderbilt University Medical Center, Nashville TN 37215, USA 
}

\maketitle  

\begin{abstract}
Integrating cross-department multi-modal data (e.g., radiology, pathology, genomic, and demographic data) is ubiquitous in brain cancer diagnosis and survival prediction. To date, such an integration is typically conducted by human physicians (and panels of experts), which can be subjective and semi-quantitative. Recent advances in multi-modal deep learning, however, have opened a door to leverage such a process in a more objective and quantitative manner. Unfortunately, the prior arts of using four modalities on brain cancer survival prediction are limited by a "complete modalities" setting (i.e., with all modalities available). Thus, there are still open questions on how to effectively predict brain cancer survival from incomplete radiology, pathology, genomic, and demographic data (e.g., one or more modalities might not be collected for a patient). For instance, should we use both complete and incomplete data, and more importantly, how do we use such data? To answer the preceding questions, we generalize the multi-modal learning on cross-department multi-modal data to a missing data setting. Our contribution is three-fold: 1) We introduce a multi-modal learning with missing data (MMD) pipeline with competitive performance and less hardware consumption; 2) We extend multi-modal learning on radiology, pathology, genomic, and demographic data into missing data scenarios; 3) A large-scale public dataset (with 962 patients) is collected to systematically evaluate glioma tumor survival prediction using four modalities. The proposed method improved the C-index of survival prediction from 0.7624 to 0.8053.

\end{abstract}

\begin{figure}[t]
\begin{center}
\includegraphics[width=0.75\linewidth]{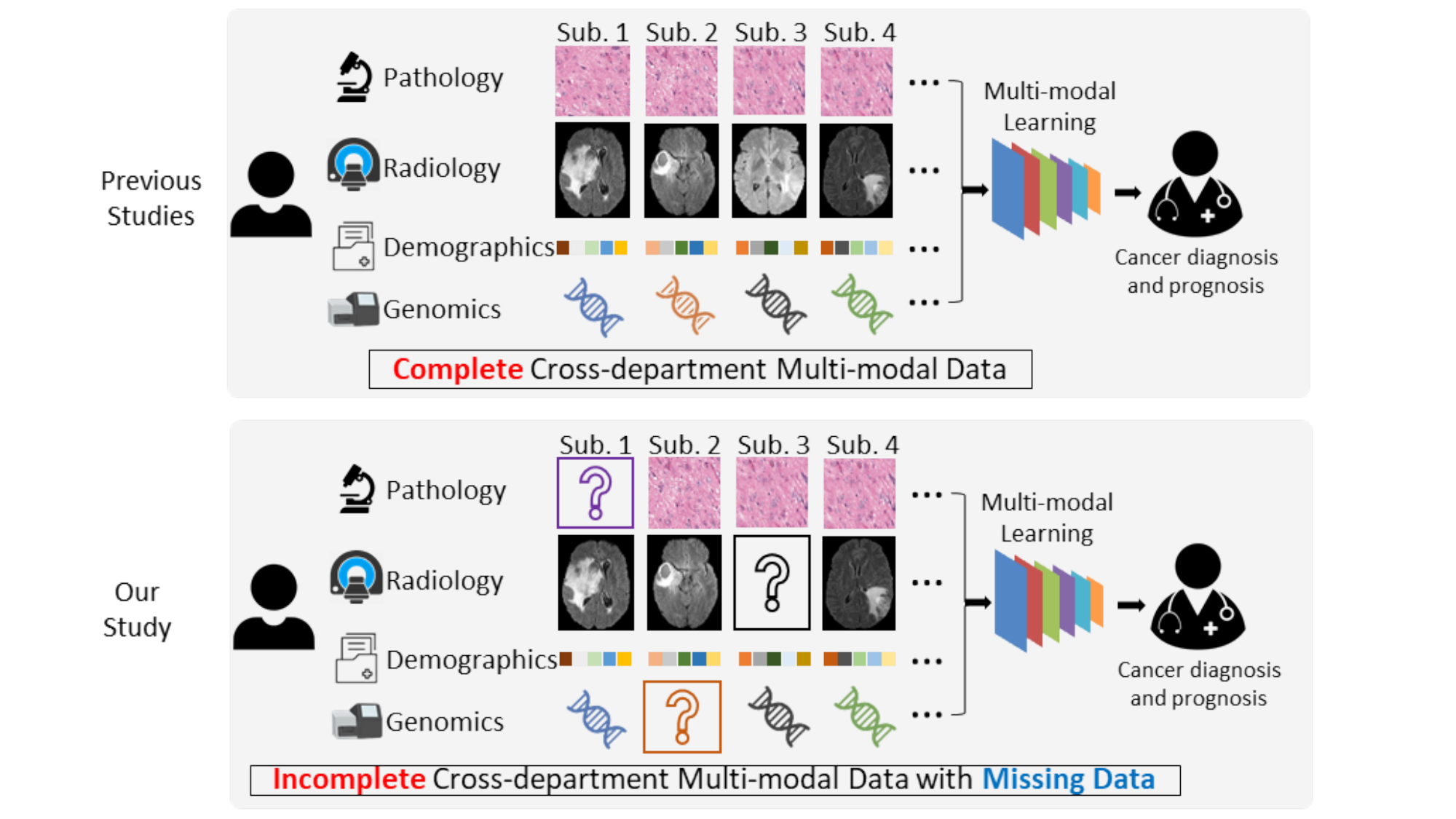}
\end{center}
   \caption{\textbf{Multi-modal learning with missing data for cancer survival prediction.} This study generalizes multi-modal learning on radiology, pathology, genomic and demographic data from a "complete data" setting (upper panel) to a "missing data" scenario (lower panel), which is ubiquitous in routine practice.}
\label{fig1:Problem Definition}
\end{figure}

\section{Introduction}
Multi-modal learning tends to improve model performance by extracting and aggregating the information of the same subject from different modalities, as compared to learning with a single modality~\cite{baltruvsaitis2018multimodal}. Medical domains are rich with multiple modalities, including image data such as pathology images, radiology images, and non-image data such as genomic data and demographic information, etc. An increasing number of works have shown that multi-modal learning helps to improve the performance of computer-aided diagnosis and prognosis in many medical applications~\cite{schneider2022integration,huang2020fusion}. 

Gliomas are the most common primary malignant brain tumors. Diagnosis and prognosis of patients with this kind of tumor are beneficial for treatment decisions and patient management ~\cite{pereira2016brain}. Some prior studies have already shown the prognostic advances of utilizing pathology images ~\cite{louis20162016}, MRI images ~\cite{bae2018radiomic}, genomic~\cite{beig2020radiogenomic} and clinical features~\cite{beig2021sexually} for the survival time prediction of glioma tumors. More recent studies have shown the promising results of integrating multi-modal data for glioma tumor prognosis~\cite{mobadersany2018predicting,zadeh2017tensor,chen2020pathomic,braman2021deep}. Mobadersany et al~\cite{mobadersany2018predicting}. concatenated genomic biomarkers and learned pathology image features. Chen et al~\cite{chen2020pathomic}. proposed to aggregate features from genomic data, pathology images and cell graphs with tensor fusion ~\cite{zadeh2017tensor}. More recently, Braman et al~\cite{braman2021deep}. extended tensor fusion with orthogonal regularization to four modalities for more precise survival prediction.

However, the above studies only utilized the comprehensive radiology, pathology, genomic, and demographic data as a "complete data" setting (all modalities are available for every patient). However, complete data are not always guaranteed in routine practice patients (Fig.~\ref{fig1:Problem Definition}). Discarding patient data with missing modalities (1) extremely aggravates the data scarcity problem of medical applications and increases the risk of overfitting~\cite{bach2017breaking}, and (2) limits the usage of the model trained by complete modalities in the inference phase because the prediction for data with complete modalities cannot be guaranteed. Cheerla et al~\cite{cheerla2019deep}. used modality dropout to improve pan-cancer classification when the dataset had missing modalities. Ghosal et al~\cite{ghosal2021g}. reconstructed uni-modal features from the multi-modal representation to force the fusion model to keep all the information of uni-modalities. However, these methods did not leverage the available modalities for the modality missing problem, and they did not thoroughly compare the model performance between the training and testing data with and without modality missing, respectively.  

\begin{figure}[t]
\begin{center}
\includegraphics[width=0.75\linewidth]{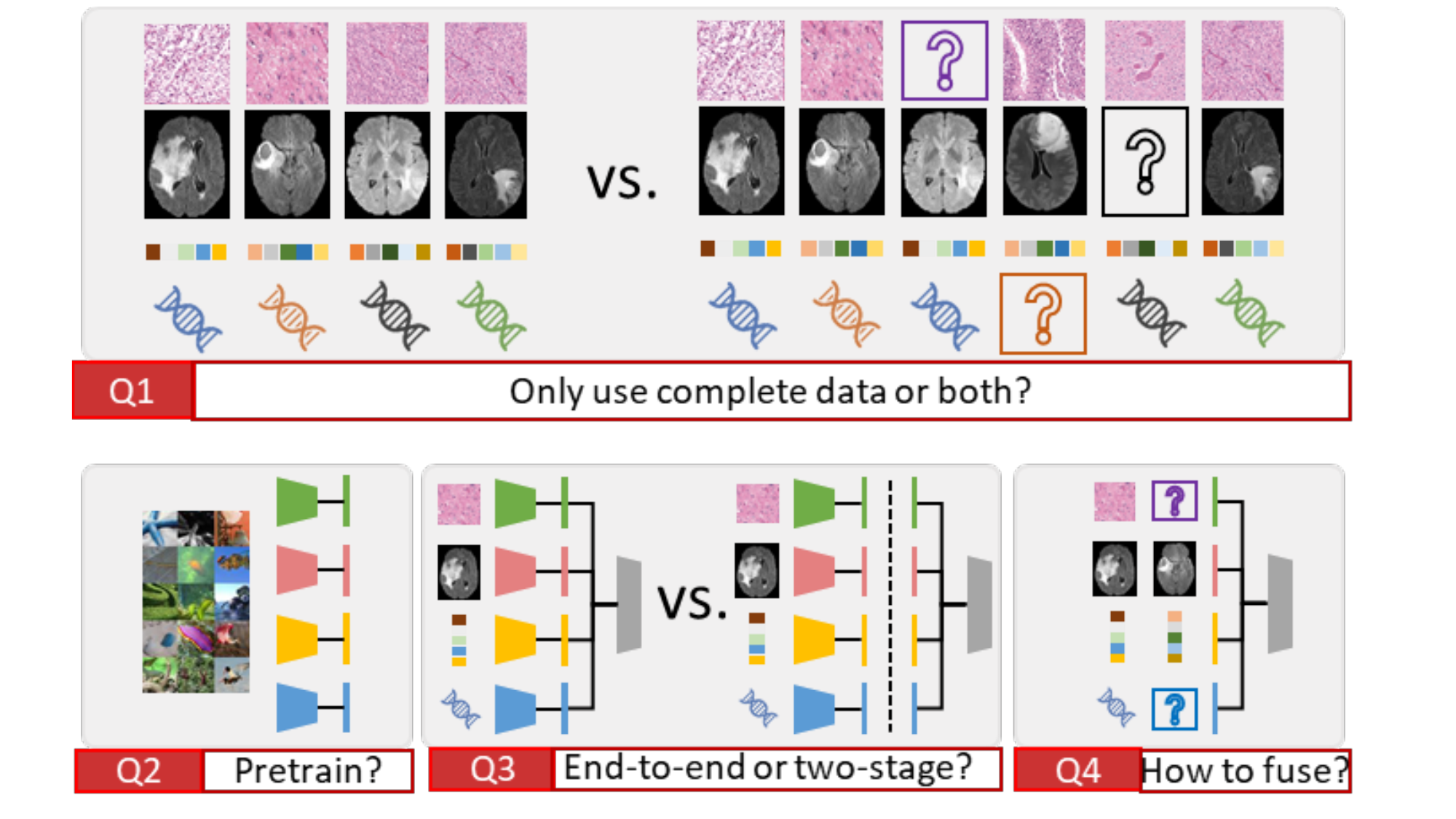}
\end{center}
   \caption{\textbf{Four questions to answer for this paper.} The experimental answers of the four questions are presented in §5.}
\label{fig2:Questions}
 \end{figure}

Therefore, there are still open questions on how to effectively learn from incomplete radiology, pathology, genomic, and demographic data (e.g., one or more modalities might not be collected for a patient) (Fig.~\ref{fig2:Questions}). Specifically, 1) \textit {Would more training data with incomplete modalities enhance the accuracy of prognosis prediction?} 2) \textit {How can we effectively utilize the pre-trained feature extractors?} 3) \textit {Which training strategy should we use, end-to-end or two-stage?} 4) \textit {How should we fuse the multi-modal information under the missing data scenarios?} 
 
 In this paper, we propose an effective multi-modal learning with missing data (MMD) pipeline while addressing the above four key questions with comprehensive experimental analyses. The contribution of this paper is three-fold:

$\bullet$ The proposed MMD method is designed for using any combinations of available data as a two-stage plug-and-play method with computational efficiency.

$\bullet$ This study generalizes multi-modal learning on radiology, pathology, genomic, and demographic data into more clinically applicable missing data scenarios.

$\bullet$ The methodological development and data analyses are performed on a large-scale dataset by combining the glioma tumor data in TCGA~\cite{TCGA-LGG, TCGA-GBM}, TCIA~\cite{clark2013cancer}, and BraTs dataset ~\cite{bakas2018identifying,bakas2017advancing,menze2014multimodal}.

\begin{figure}[t]
\begin{center}
\includegraphics[width=0.80\linewidth]{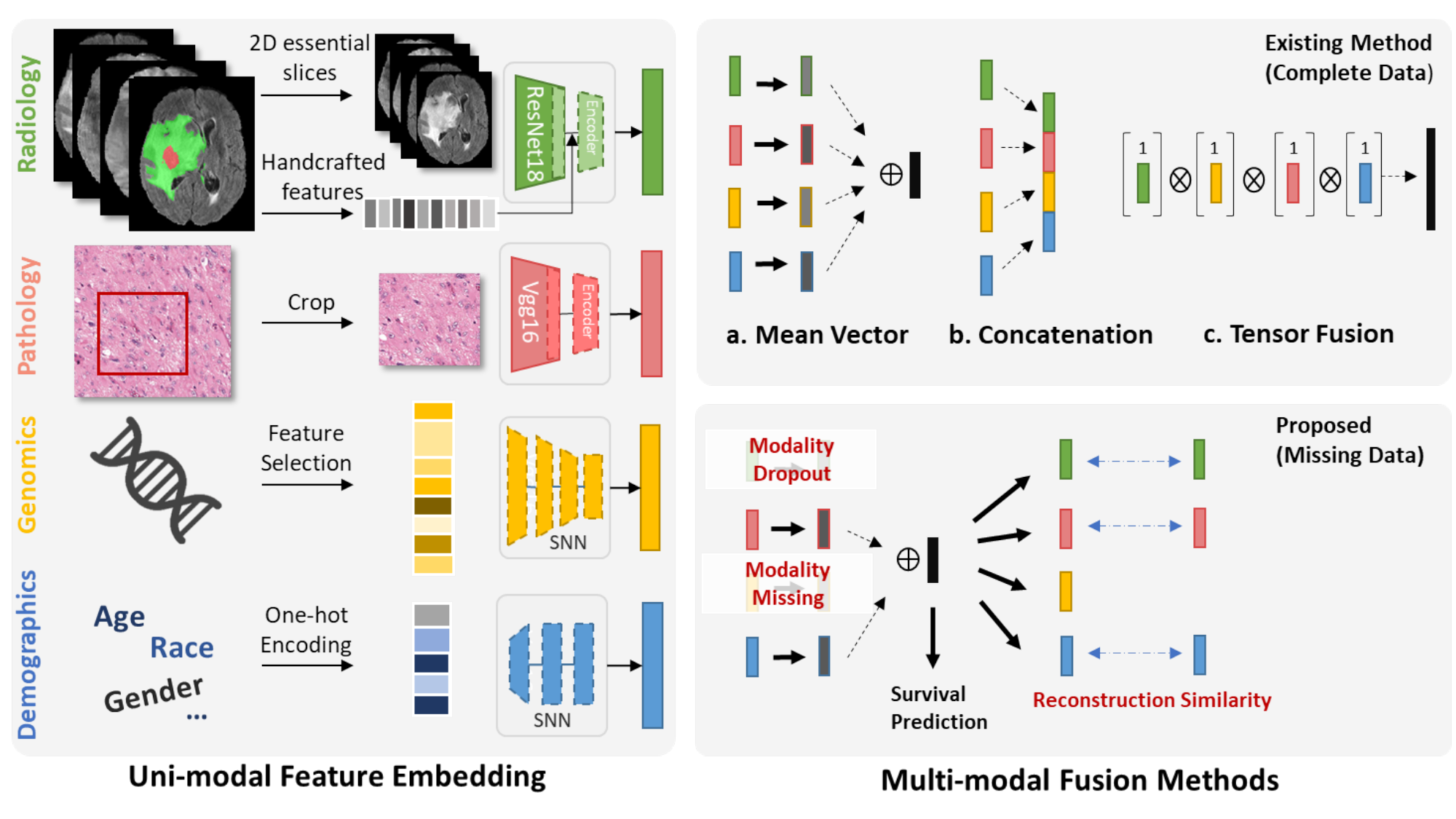}
\end{center}
   \caption{\textbf{The pipeline of proposed multi-modal learning model with missing modalities.} Uni-modal deep neural networks are used to generate features of different modalities (left panel). Different fusion methods are shown in the right panel.}
\label{fig3:Pipeline}
 \end{figure}

\section{Methods}

The multi-modal fusion learning approach of the proposed MMD method, as well as the methods in prior arts, are presented in Fig.~\ref{fig3:Pipeline}. The MMD pipeline is a two-stage framework that is designed for missing data scenarios, including the (1) uni-modal feature embedding, and (2) multi-modal information fusion.

\subsection{Uni-modal feature embedding}
Low-dimensional features are obtained for each modality by training independent deep survival models using different neural networks. 

\textbf{Radiology Modal.} 
For registered 3D radiology MRI scans, the tumor and edema regions are automatically obtained by a trained nnUNet~\cite{isensee2019nnu} segmentation model. The 2D essential slices (resized to 120 x 120) with the largest tumor and edema region from the four modalities are stacked as a four-channel input. To bridge the inconsistency between a pre-trained model and the new input dimensionality, the weights of the $1^{st}$ channel of a pre-trained ResNet-18 are copied to the $4^{th}$ channel as the weight initialization. In addition, 318 2D and 3D handcrafted features are extracted by PyRadiomics from tumor volumes to provide extra shape, location, and texture information. The handcrafted features are concatenated to the image features learned by the ResNet-18. Based on our experiments, adding handcrafted features improve the model performance as compared to using CNN features along.   

\textbf{Pathology Modal.} 
Following the prior work~\cite{chen2020pathomic}, ImageNet pre-trained Vgg-16 is used to extract features from pathology images. Same as the radiology modality, only the last layers of the pre-trained network (ResNet and Vgg) is finetuned, as opposed to retraining the entire network. Each pathology image with the size of 1024 x 1024 is extracted from a tumor region ~\cite{mobadersany2018predicting, chen2020pathomic}. An image patch with the size of 512 x 512 is extracted from each image in each training iteration for learning. In the testing phase, a sliding window with the size of 512 x 512 and a stride of 256 is used to crop each image into 9 patches. The average aggregation of the prediction of these 9 patches is the prediction for each pathology image sample. 

\textbf{Genomics and Demographics Modals.} 
The genomic modality consists of 80 DNA features including 79 of the most expressive features from copy number variations (CNV) and one feature from mutation status  (binary indication of mutation status for IDH1 gene, 0/1), same as the previous work \cite{braman2021deep}.  For the demographic modality, we use four features including age, gender, race and ethnicity. They are extended to 9 features through applying one-hot encoding on categorical features. Following the prior work~\cite{chen2020pathomic}, self-normalized networks ~\cite{klambauer2017self} are employed for both genomic data and demographic data to mitigate overfitting. 

\subsection{Multi-modal fusion with complete data}
In prior arts, three canonical fusion strategies have been broadly used on the complete datasets, including (1) feature concatenation ~\cite{yap2018multimodal, mobadersany2018predicting}, (2) mean vector (i.e. element-wise average) ~\cite{ghosal2021g, cheerla2019deep}, and (3) Kronecker product based tensor fusion ~\cite{chen2020pathomic, wang2021gpdbn}. In our study, such fusion methods are used to fuse the features, followed by a three-layer multi-layer perceptron (MLP) with ReLU activations and dropout layers for survival prediction. For the \textbf{concatenation method}, the 1$\times$32 feature embeddings of each modality are concatenated directly to generate a multi-modal representation with a length of 128. For the \textbf{mean vector fusion}, the uni-modal embeddings are firstly extended to 128 dimensional vectors with a modal-specific two-layer perceptron, and a reverse network structure is used to reconstruct the uni-modal embeddings from the mean vector. For the \textbf{pathomic fusion method}, we directly extend the tensor fusion with gated attention method proposed by Chen et al~\cite{chen2020pathomic}. from three to four uni-modal 1 $\times$ 32 embeddings. 

\subsection{Multi-modal fusion for missing data}
In the missing data setting, we extend multi-modal fusion with the MMD method, which fuses the feature embeddings from all available modalities. Among the above three canonical fusion strategies, we employ the mean vector (element-wise average of available embeddings) as the backbone fusion method since its "order-less" nature leads to larger flexibility in using missing data. To further boost fusion capability on missing data, we introduce modality dropout~\cite{neverova2015moddrop} by randomly discarding modalities for training. Only the rest of uni-modal embeddings are calculated for the mean vector. 

To encourage the model to learn cross-modality information, modality reconstruction is used in the MMD method to fit a complete multi-modal representation from all available modalities. Moreover, different from previous work ~\cite{ghosal2021g}, for the modalities that are intentionally dropped out, we ask the MMD model to reconstruct such modalities as well, since the original modalities are actually known. 

%\begin{equation}
%    \mathcal x_{mean}^{i} = \frac{1}{\left | {\alpha_{i}}=1 %\right|} (\sum_v^V {\alpha_{i}^{v} x_{i}^{v}}),
%\end{equation}

The loss function to supervise the proposed method consists of two terms. The first term is a cox loss that is defined as: 
\begin{equation}
    \mathcal L_{cox} = - \sum_{i: E_{i} =1}({ F_{\theta}( h_{i} ) - log  \sum_{j: t_{i}  	\geq t_{j} }{ e^{ F_{\theta}( h_{j} ) } } }),
\end{equation}
where the values $t_{i}$, $E_{i}$, $h_{i}$ are survival time, censor status and multi-modal representations of sample $i$. If the death is observed, censor status $E_{i} = 1$. $F_{\theta}$ denotes the networks which predict the survival hazard (risk of death) from $h_{i}$. The second term is the reconstruction loss:
\begin{equation}
    \mathcal L_{recon} = \frac{1}{\sum_{i}^{N}{\left | {\alpha_{i}}=1 \right|} }\sum_{i}^{N}(   \sum_v^V {\alpha_{i}^{v} \parallel x_{i}^{v},        \widetilde{x}_{i}^{v}\parallel}),
\end{equation}
where $x_{i}^{v}$ is the embedding of modality $v \in V$ of sample $i \in N$. $\widetilde{x}_{i}^{v}$ is the reconstructed embedding of modality $v$ and $\alpha_{i}^{v}$ indicates the original availability of the modality $v$ of subject $i$. $\alpha_{i}^{v} = 0$ when this modality is missing and ${\left| {\alpha_{i}}>0\right|}$ represents the number of available modalities for sample $i$. $L_{2}$ $norm$ is used to calculate the difference between the reconstructed uni-modal embeddings and the corresponding real uni-modal embeddings. 
The overall loss becomes:
\begin{equation}
    \mathcal L_{total}= \mathcal L_{cox}+\lambda \mathcal L_{recon}
\end{equation} 
where a weight $\lambda = 1$ is empirically selected to balance two losses. 

\begin{figure}[t]
\begin{center}
\includegraphics[width=0.70\linewidth]{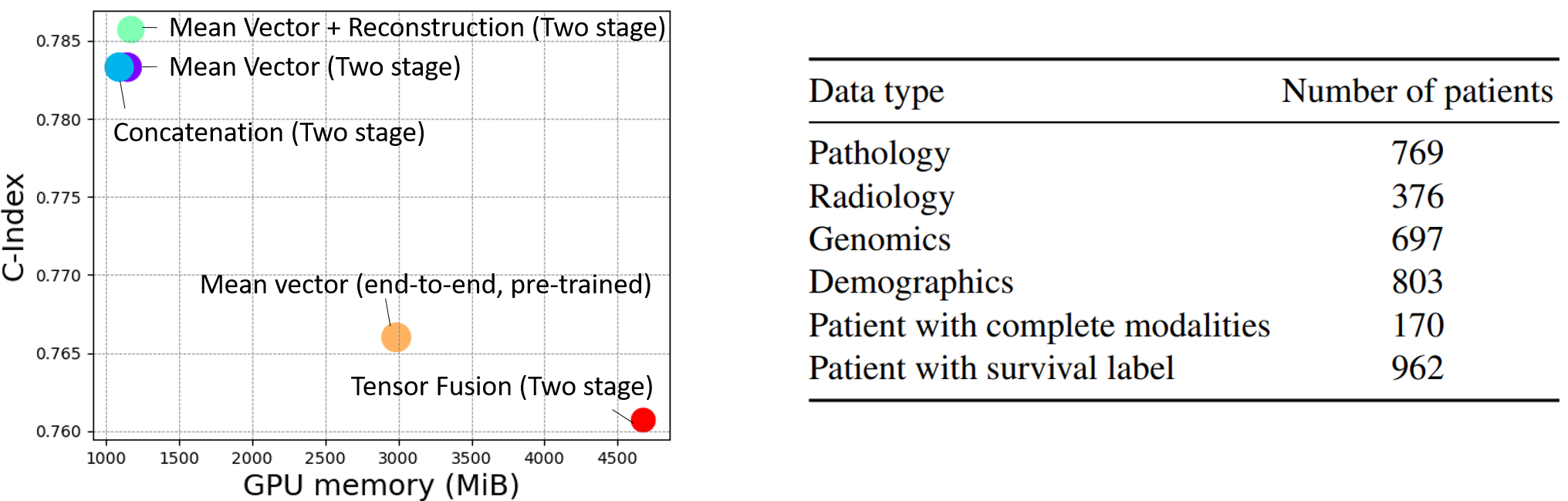}
\end{center}
   \caption{The left panel shows the GPU memory consumption and c-index of different methods (with the best performance). The right panel shows the number of patients in different modalities.}
\label{fig4:Supplementary Results}
 \end{figure}

\section{Data and Experimental Setting}
We collected a large-scale dataset with both complete and incomplete glioma tumor data (1698 samples from 962 patients) by combining TCGA~\cite{TCGA-LGG,TCGA-GBM}, TCIA~\cite{clark2013cancer}, and the BraTs dataset~\cite{bakas2018identifying,bakas2017advancing,menze2014multimodal}. Each patient had at least one modality available. All patients had survival time and censor status. The detailed data information are presented in 
Fig.~\ref{fig4:Supplementary Results}. Following the train-test splits of previous works ~\cite{chen2020pathomic, mobadersany2018predicting}, we enlarged the training set with more patients from the BraTs dataset and used the subset of original testing sets with four modalities available. Also, 10\% of patients in training set were randomly selected for validation set. In the routine clinical situation, the acquisition of pathology and genomic modalities was more invasive to achieve than the non-invasive radiology and demographic data. Thus, we also evaluated scenarios in which pathology and genomic data were missing (Table~\ref{table:performance3}). Each modality was randomly dropped out (set to zero) by an empirically selected rate 0.5 until only one modality was left for a sample. For the end-to-end fusion strategy, the network weights pre-trained by ImageNet or the uni-modal data were loaded. Then, the uni-modal models and the fusion network were trained together. As for the two-stage strategy, 32-dimensional uni-modal embeddings prepared by trained uni-modal models were used as inputs to train the fusion network only. Specifically, a 512 x 512 patch was randomly cropped from each pathology image for an embedding in training. To evaluate the survival prediction, the concordance index (c-index) value was used. The average aggregation of multiple patches from multiple samples of each patient is calculated as the final prediction. Intensity normalized images were augmented by rotation, flipping, and color jitter methods in training. The batch size was 32 for non-image uni-modal training and 8 for image training, with a learning rate of 0.002. The batch size of the fusion was set to 8 with a learning rate of 0.0002. All the experiments were run on a 16GB NVIDIA GPU. The code is available at: https://github.com/cuicathy/MMD\_SurvivalPrediction.git

\section{Results}
Table~\ref{table:performance} compared our proposed pipeline with prior benchmarks~\cite{chen2020pathomic,braman2021deep}. Our proposed MMD method achieved superior performance compared with benchmarks. The performance on joint testing sets of pathomic fusion~\cite{chen2020pathomic} and our work are used for a fair comparison of different methods. The best performance of their released model (the "pathgraphomic" network with attention gates trained by mRNA, DNA and pathology images) was compared with our best model (all data + modality dropout + reconstruction + mean vector) on this subset. We rebuilt the Deep orthogonal fusion following the papers \cite{braman2021deep, lezama2018ole} (e.g., uni-modal embeddings were compressed to 8 dimension by a linear layer for tensor fusion), but the four uni-modal structures and features mentioned above trained by all available uni-modal data were used for a fair comparison.

\begin{table}[t]
\caption{Survivals prediction performance with benchmarks}
\centering
\begin{tabular}{p{4cm}p{3cm}p{3cm}}
\toprule
 Methods & C-Index \\
\midrule
Pathomic Fusion \cite{chen2020pathomic} & 0.7697 ± 0.047 \\
Deep Orthogonal Fusion \cite{braman2021deep} &  0.7624 ± 0.042 \\ 
MMD (Ours) & \textbf{0.8053 ± 0.038}  \\
\bottomrule

\label{table:performance}
\end{tabular}

\caption{Survival prediction results of different uni-modal model training strategies and fusion strategies using all data or complete data. "C" means training using complete data while "M" means data with missing modalities. Same layers were froze in the training of multi-modal fusion as the uni-modal modals. * Only the weights pre-trained by ImageNet were loaded. ** Weights of uni-modal networks were loaded for finetuning.}
\centering
\begin{tabular}{p{4cm}p{3.5cm}p{2cm}p{3cm}}
\toprule
Modality & Training Strategy &Training Data & C-index \\
\midrule
Uni-modal (Radiology) & finetune entire network & C+M &  0.6957 ± 0.043 \\
Uni-modal (Pathology) & finetune entire network & C+M &  0.6803 ± 0.008 \\
Uni-modal (Radiology) & finetune last layers & C+M & \textbf{0.7062 ± 0.039} \\
Uni-modal (Pathology) & finetune last layers & C+M & \textbf{0.7319 ± 0.026} \\
\midrule
Multi-modal (Mean vector) & end-to-end (from scratch*) & C & 0.7263 ± 0.027 \\
Multi-modal (Mean vector)& end-to-end (from scratch*)& C+M & 0.7609 ± 0.016\\
Multi-modal (Mean vector)& end-to-end (finetune**) & C & 0.7580 ± 0.030 \\ 
Multi-modal (Mean vector)& end-to-end (finetune**) & C+M& 0.7607 ± 0.021\\

Multi-modal (Mean vector)& two-stage & C & 0.7571 ± 0.034\\ 
Multi-modal (Mean vector)& two-stage & C+M &\textbf{0.7717 ± 0.034}\\
\bottomrule
\label{table:performance2}
\end{tabular}
\end{table}
Results of different training strategies for uni-modal learning and multi-modal fusion are in Table~\ref{table:performance2}. For uni-modal training, only finetuning the last layer of the ImageNet pre-trained network achieved better uni-modal survival prediction than finetuning the whole pre-trained network. When using multi-modal data, two-stage training with fixed uni-modal embeddings is competitive compared with the end-to-end training. Table~\ref{table:performance3} shows  the results of three different multi-modal fusion methods trained by complete data (``C" in the table) or all data ("C+M" in the table), with or without modality dropout. For the mean vector method, the modality dropout, reconstruction, and training with all data consistently improved the performance on testing sets.     

\begin{table}[t]
\caption{Survival prediction results (C-index) of different training strategies}
\begin{tabular}{|clllllll|}
\hline
\rowcolor[HTML]{C0C0C0} 
\multicolumn{5}{|c|}{\cellcolor[HTML]{C0C0C0}Training} &
  \multicolumn{3}{c|}{\cellcolor[HTML]{C0C0C0}Testing} \\ \hline
\rowcolor[HTML]{EFEFEF} 
\multicolumn{1}{|l|}{\cellcolor[HTML]{EFEFEF}\begin{tabular}[c]{@{}l@{}}Fusion \\ method\end{tabular}} &
  \multicolumn{1}{l|}{\cellcolor[HTML]{EFEFEF}\begin{tabular}[c]{@{}l@{}}Data for \\ uni-modal \\ embedding\end{tabular}} &
  \multicolumn{1}{l|}{\cellcolor[HTML]{EFEFEF}\begin{tabular}[c]{@{}l@{}}Data for \\ multi-modal \\ fusion\end{tabular}} &
  \multicolumn{1}{l|}{\cellcolor[HTML]{EFEFEF}\begin{tabular}[c]{@{}l@{}}Drop-\\ out\end{tabular}} &
  \multicolumn{1}{l|}{\cellcolor[HTML]{EFEFEF}Recon} &
  \multicolumn{1}{c|}{\cellcolor[HTML]{EFEFEF}\begin{tabular}[c]{@{}c@{}}Complete \\ modalities\end{tabular}} &
  \multicolumn{1}{c|}{\cellcolor[HTML]{EFEFEF}\begin{tabular}[c]{@{}c@{}}Pathology \\ missing\end{tabular}} &
  \multicolumn{1}{c|}{\cellcolor[HTML]{EFEFEF}\begin{tabular}[c]{@{}c@{}}Gene and\\  pathology \\ missing\end{tabular}} \\ \hline
\multicolumn{1}{|c|}{} &
  \multicolumn{1}{l|}{C+M} &
  \multicolumn{1}{l|}{C} &
  \multicolumn{1}{l|}{} &
  \multicolumn{1}{l|}{} &
  \multicolumn{1}{l|}{0.7659 ± 0.032} &
  \multicolumn{1}{l|}{0.7563 ± 0.035} &
  0.7169 ± 0.035 \\ \cline{2-8} 
\multicolumn{1}{|c|}{} &
  \multicolumn{1}{l|}{C+M} &
  \multicolumn{1}{l|}{C+M} &
  \multicolumn{1}{l|}{} &
  \multicolumn{1}{l|}{} &
  \multicolumn{1}{l|}{0.7743 ± 0.034} &
  \multicolumn{1}{l|}{0.7647 ± 0.030} &
  0.7278 ± 0.037 \\ \cline{2-8} 
\multicolumn{1}{|c|}{} &
  \multicolumn{1}{l|}{C+M} &
  \multicolumn{1}{l|}{C} &
  \multicolumn{1}{l|}{\checkmark} &
  \multicolumn{1}{l|}{} &
  \multicolumn{1}{l|}{\underline{0.7833 ± 0.030}} &
  \multicolumn{1}{l|}{\underline{0.7725 ± 0.031}} &
  0.7275 ± 0.037 \\ \cline{2-8} 
\multicolumn{1}{|c|}{\multirow{-4}{*}{\begin{tabular}[c]{@{}c@{}}Concat-\\ enation\end{tabular}}} &
  \multicolumn{1}{l|}{C+M} &
  \multicolumn{1}{l|}{C+M} &
  \multicolumn{1}{l|}{\checkmark} &
  \multicolumn{1}{l|}{} &
  \multicolumn{1}{l|}{0.7817 ± 0.029} &
  \multicolumn{1}{l|}{0.7668 ± 0.032} &
  {\underline{0.7352 ± 0.034}} \\ \hline
\multicolumn{8}{|l|}{} \\ \hline
\multicolumn{1}{|l|}{} &
  \multicolumn{1}{l|}{C+M} &
  \multicolumn{1}{l|}{C} &
  \multicolumn{1}{l|}{} &
  \multicolumn{1}{l|}{} &
  \multicolumn{1}{l|}{0.7503 ± 0.032} &
  \multicolumn{1}{l|}{0.7274 ± 0.030} &
  0.6926 ± 0.045 \\ \cline{2-8} 
\multicolumn{1}{|l|}{} &
  \multicolumn{1}{l|}{C+M} &
  \multicolumn{1}{l|}{C+M} &
  \multicolumn{1}{l|}{} &
  \multicolumn{1}{l|}{} &
  \multicolumn{1}{l|}{0.7483 ± 0.026} &
  \multicolumn{1}{l|}{0.7328 ± 0.036} &
  {0.6926 ± 0.044} \\ \cline{2-8} 
\multicolumn{1}{|l|}{} &
  \multicolumn{1}{l|}{C+M} &
  \multicolumn{1}{l|}{C} &
  \multicolumn{1}{l|}{\checkmark} &
  \multicolumn{1}{l|}{} &
  \multicolumn{1}{l|}{0.7513 ± 0.035} &
  \multicolumn{1}{l|}{\underline{0.7386 ± 0.035}} &
  \underline{0.7187 ± 0.040} \\ \cline{2-8} 
\multicolumn{1}{|l|}{\multirow{-4}{*}{\begin{tabular}[c]{@{}l@{}}Tensor \\ Fusion\end{tabular}}} &
  \multicolumn{1}{l|}{C+M} &
  \multicolumn{1}{l|}{C+M} &
  \multicolumn{1}{l|}{\checkmark} &
  \multicolumn{1}{l|}{} &
  \multicolumn{1}{l|}{\underline{0.7660 ± 0.031}} &
  \multicolumn{1}{l|}{0.7314 ± 0.034} &
  {0.7117 ± 0.035} \\ \hline
\multicolumn{8}{|l|}{} \\ \hline
\multicolumn{1}{|c|}{} &
  \multicolumn{1}{l|}{C} &
  \multicolumn{1}{l|}{C} &
  \multicolumn{1}{l|}{} &
  \multicolumn{1}{l|}{} &
  \multicolumn{1}{l|}{0.7597 ± 0.029} &
  \multicolumn{1}{l|}{0.7464 ± 0.032} &
  0.7013 ± 0.038 \\ \cline{2-8} 
\multicolumn{1}{|c|}{} &
  \multicolumn{1}{l|}{C+M} &
  \multicolumn{1}{l|}{C} &
  \multicolumn{1}{l|}{} &
  \multicolumn{1}{l|}{} &
  \multicolumn{1}{l|}{0.7571 ± 0.034} &
  \multicolumn{1}{l|}{0.7495 ± 0.036} &
  0.7118 ± 0.037 \\ \cline{2-8} 
\multicolumn{1}{|c|}{} &
  \multicolumn{1}{l|}{C+M} &
  \multicolumn{1}{l|}{C+M} &
  \multicolumn{1}{l|}{} &
  \multicolumn{1}{l|}{} &
  \multicolumn{1}{l|}{0.7717 ± 0.034} &
  \multicolumn{1}{l|}{0.7622 ± 0.034} &
  0.7321 ± 0.036 \\ \cline{2-8} 
\multicolumn{1}{|c|}{} &
  \multicolumn{1}{l|}{C} &
  \multicolumn{1}{l|}{C} &
  \multicolumn{1}{l|}{\checkmark} &
  \multicolumn{1}{l|}{} &
  \multicolumn{1}{l|}{0.7740 ± 0.026} &
  \multicolumn{1}{l|}{0.7644 ± 0.029} &
  0.7081 ± 0.032 \\ \cline{2-8} 
\multicolumn{1}{|c|}{} &
  \multicolumn{1}{l|}{C+M} &
  \multicolumn{1}{l|}{C} &
  \multicolumn{1}{l|}{\checkmark} &
  \multicolumn{1}{l|}{} &
  \multicolumn{1}{l|}{0.7821 ± 0.029} &
  \multicolumn{1}{l|}{0.7683 ± 0.032} &
  0.7223 ± 0.037 \\ \cline{2-8} 
\multicolumn{1}{|c|}{} &
  \multicolumn{1}{l|}{C+M} &
  \multicolumn{1}{l|}{C+M} &
  \multicolumn{1}{l|}{\checkmark} &
  \multicolumn{1}{l|}{} &
  \multicolumn{1}{l|}{\underline{0.7833 ± 0.030}} &
  \multicolumn{1}{l|}{\underline{0.7702 ± 0.032}} &
  {\underline{0.7373 ± 0.033}} \\ \cline{2-8} 
\multicolumn{1}{|c|}{} &
  \multicolumn{1}{l|}{C+M} &
  \multicolumn{1}{l|}{C} &
  \multicolumn{1}{l|}{} &
  \multicolumn{1}{l|}{\checkmark} &
  \multicolumn{1}{l|}{0.7660 ±  0.029} &
  \multicolumn{1}{l|}{0.7576 ± 0.032} &
  0.7131 ± 0.034 \\ \cline{2-8} 
\multicolumn{1}{|c|}{} &
  \multicolumn{1}{l|}{C+M} &
  \multicolumn{1}{l|}{C+M} &
  \multicolumn{1}{l|}{} &
  \multicolumn{1}{l|}{\checkmark} &
  \multicolumn{1}{l|}{0.7779 ±  0.032} &
  \multicolumn{1}{l|}{0.7708 ± 0.033} &
  0.7255 ± 0.037 \\ \cline{2-8} 
\multicolumn{1}{|c|}{} &
  \multicolumn{1}{l|}{C+M} &
  \multicolumn{1}{l|}{C} &
  \multicolumn{1}{l|}{\checkmark} &
  \multicolumn{1}{l|}{\checkmark} &
  \multicolumn{1}{l|}{0.7812 ± 0.027} &
  \multicolumn{1}{l|}{0.7609 ± 0.030} &
  0.7164 ± 0.033 \\ \cline{2-8} 
\multicolumn{1}{|c|}{\multirow{-10}{*}{\begin{tabular}[c]{@{}c@{}}Mean   \\ Vector\end{tabular}}} &
  \multicolumn{1}{l|}{C+M} &
  \multicolumn{1}{l|}{C+M} &
  \multicolumn{1}{l|}{\checkmark} &
  \multicolumn{1}{l|}{\checkmark} &
  \multicolumn{1}{l|}{\textbf{0.7857 ± 0.026}} &
  \multicolumn{1}{l|}{\textbf{0.7808 ± 0.032}} &
  {\textbf{0.7451 ± 0.034}} \\ \hline
\end{tabular}
{* ``C" means training using complete data while ``M" means missing data. The best performances among all settings are highlighted in \textbf{bold}, while the best performances of a certain fusion method (without reconstruction loss) are highlighted by \underline{underline}.}
\label{table:performance3}
\end{table}

\section{Ablation Studies for Answering Four Questions}
To answer the four questions in Fig.~\ref{fig2:Questions}, we conducted comprehensive studies as shown in Table.~\ref{table:performance2} and~\ref{table:performance3}.

\textbf{Q1: Would more training data with incomplete modalities enhance the accuracy of prognosis prediction?}
Table~\ref{table:performance2} and~\ref{table:performance3} results indicated that using both complete and incomplete modalities can improve the c-index for survival prediction.

\textbf{Q2: How do we effectively utilize the pre-trained feature extractors?}
Fig.~\ref{fig4:Supplementary Results} and Table~\ref{table:performance2} indicated that only finetuning the last layer yields better performance compared with finetuning the entire network in our task. 

\textbf{Q3: Which training strategy should we use, end-to-end or two-stage?}
Table~\ref{table:performance2} indicated that the two-stage learning strategy yielded a competitive performance compared with the end-to-end counterparts, and it is memory efficient for more flexibility.

\textbf{Q4: How should we fuse multi-modal information under the missing data scenarios?}
The comprehensive ablation studies of different fusion strategies have been provided in Table~\ref{table:performance3}. The proposed MMD approach achieved superior performance among different strategies.

\section{Conclusion}
In this paper, we generalize the cross-department multi-modal data (radiology, pathology, genomic, and demographic data) deep survival learning from a complete data setting to the more clinically applicable missing data scenarios. First, we presented the MMD method, a multi-modal prognosis pipeline with an effective and efficient design of using both modality complete and incomplete data. Second, with a relatively large-scale cohort, we investigated four key questions of utilizing modality missing data for the brain cancer survival analysis. Future work might include applying this method on external datasets and exploring other uni-modal embedding methods. We hope this study could also be used as a reference for future studies in predicting brain cancer prognosis.

\section{Acknowledgements}
This work is supported by the Leona M. and Harry B. Helmsley Charitable Trust grant G-1903-03793, NSF CAREER 1452485. This work is in part based upon data generated by the TCGA Research Network: https://www-cancer-gov.proxy.library.vanderbilt.edu/tcga.

\bibliographystyle{splncs04}
\bibliography{main}
\end{document}